\documentclass[10pt]{article}

\usepackage[T1]{fontenc}
\usepackage{times}
\usepackage{microtype}
\usepackage{graphicx}
\usepackage{subcaption}
\usepackage{booktabs}
\usepackage{makecell}
\usepackage{placeins}
\usepackage{hyperref}
\usepackage{algorithm}
\usepackage{algorithmic}
\usepackage{amsmath}
\usepackage{amssymb}
\usepackage{mathtools}
\usepackage{amsthm}
\usepackage[capitalize,noabbrev]{cleveref}
\usepackage{natbib}
\usepackage{fancyhdr}

\makeatletter

\def\section{\@startsection{section}{1}{\z@}{-0.12in}{0.02in}{\large\bfseries\raggedright}}
\def\subsection{\@startsection{subsection}{2}{\z@}{-0.10in}{0.01in}{\normalsize\bfseries\raggedright}}
\def\subsubsection{\@startsection{subsubsection}{3}{\z@}{-0.08in}{0.01in}{\normalsize\bfseries\itshape\raggedright}}
\makeatother

\newcommand{\PreprintTitle}[1]{
  \thispagestyle{plain}
  {\center\baselineskip 18pt
	\hrule height1pt\vskip .25in
	{\Large\bfseries #1\par}
	\vskip .22in
	\hrule height1pt\vskip .3in}
}

\theoremstyle{plain}

\theoremstyle{definition}

\theoremstyle{remark}

\begin{document}

\twocolumn[
\PreprintTitle{DT-Transformer: A Foundation Model for Disease Trajectory Prediction on a Real-world Health System}

\begin{center}
\begin{minipage}{0.92\textwidth}
\raggedright
{\normalsize
\begin{center}
Yunying Zhu\textsuperscript{1,2,*} \quad
Andrew R Weckstein\textsuperscript{1,2,*} \quad
Kueiyu Joshua Lin\textsuperscript{1} \quad
Jie Yang\textsuperscript{1,3,4,5}
\end{center}
\vspace{0.05in}
\textsuperscript{1}Division of Pharmacoepidemiology and Pharmacoeconomics, Department of Medicine, Brigham and Women's Hospital, Harvard Medical School, Boston, MA, USA\\
\textsuperscript{2}Harvard T.H. Chan School of Public Health, Harvard University, Boston, MA, USA\\
\textsuperscript{3}Kempner Institute for the Study of Natural and Artificial Intelligence, Harvard University, MA, USA\\
\textsuperscript{4}Broad Institute of MIT and Harvard, Cambridge, MA, USA\\
\textsuperscript{5}Harvard Data Science Initiative, Harvard University, Cambridge, MA, USA\\
\vspace{0.03in}
{\bfseries \textsuperscript{*}These authors contributed equally to the preparation of this manuscript.}\\
\textbf{Correspondence:}
Jie Yang, PhD (\texttt{jyang66@bwh.harvard.edu}), Division of Pharmacoepidemiology and Pharmacoeconomics,
Brigham and Women's Hospital \& Harvard Medical School, 75 Francis St, Boston, MA 02115, USA
\vspace{0.05in}
}
\end{minipage}
\end{center}

\vspace{0.15in}
]

\begin{abstract}
Accurate disease trajectory prediction is critical for early intervention, resource allocation, and improving long-term outcomes. While electronic health records (EHRs) provide a rich longitudinal view of patient health in clinical environments, models trained on curated research cohorts may not reflect routine deployment settings, and those trained on single-hospital datasets capture only fragments of each patient's trajectory. This highlights the importance of leveraging large, multi-hospital health systems for training and validation to better reflect real-world clinical complexity. In this work, we develop DT-Transformer, a foundation model trained on 57.1M structured EHR entries over 1.7M patients from Mass General Brigham (MGB), spanning 11 hospitals and a broad network of outpatient clinics \cite{desaiBroadeningReachFDASentinel2021}. DT-Transformer achieves strong discrimination in both held-out and prospective validation settings. Next-event prediction achieves a median age- and sex-stratified AUC of 0.871 across 896 disease categories, with all categories exceeding AUC 0.5. These results support health system-scale training as a path toward foundation models suited to real-world clinical forecasting.
\end{abstract}

\section{Introduction}
Predicting how disease develops in individual patients is a long-standing goal in medicine, requiring models that can learn from longitudinal data collected over many years. Structured electronic health records (EHRs), which capture routine care for millions of patients, offer a rich data source for this task. Inspired by the success of large language models (LLMs), researchers have applied transformer-based architectures \cite{vaswaniAttentionAllYou2023} to this setting, treating patient histories as ordered sequences and learning latent representations of disease progression. These efforts range from disease-specific predictors \cite{mengBidirectionalRepresentationLearning2021, placidoDeepLearningAlgorithm2023, maoADBERTUsingPretrained2023, yangTransformEHRTransformerbasedEncoderdecoder2023, rasmyMedBERTPretrainedContextualized2021} to foundation models that predict time-to-event outcomes \cite{steinbergMOTORTIMETOEVENTFOUNDATION2024}, jointly model hundreds of diseases \cite{shmatkoLearningNaturalHistory2025}, or simulate full trajectories \cite{rencZeroShotHealth2024a, makarovLargeLanguageModels2025}. Among these, Delphi-2M introduced a generative pretrained transformer (GPT)-style architecture that models diagnosis timing across many conditions, supporting next-event prediction and autoregressive trajectory simulation \cite{shmatkoLearningNaturalHistory2025}.

Despite this progress, the data sources used to train and evaluate these models limit conclusions about their utility in routine care. Existing trajectory models have largely been developed in either curated research cohorts or single-hospital EHRs. Delphi-2M, for instance, was trained on UK Biobank records, which reflect a highly selected population \cite{fryComparisonSociodemographicHealthRelated2017} and research-oriented data collection that does not mirror routine clinical practice. Models trained on single-hospital EHRs, such as ETHOS \cite{rencZeroShotHealth2024a, johnson_mimic-iv_2023}, capture only one institution's care and miss progression that occurs elsewhere. These gaps motivate evaluation on data drawn from large, multi-setting health systems, which capture real-world trajectories across care settings and over longer horizons.

In this work, we introduce DT-Transformer, a foundation model for disease trajectory prediction adapted to longitudinal structured EHR data from Mass General Brigham (MGB). DT-Transformer is trained on 57.1 million structured diagnosis events drawn from EHRs at MGB, a large integrated health system encompassing 11 hospitals and hundreds of affiliate clinics \cite{desaiBroadeningReachFDASentinel2021}. DT-Transformer achieves strong disease prediction across nearly 900 diseases, demonstrating the transferability of transformer frameworks to a real-world EHR setting. We find that near-term discrimination attenuates more rapidly than for models trained on sparser biobank-style data, likely reflecting structural differences in data density and observation gaps between research and real-world settings. We also find that expanding training to include richer diagnostic histories did not improve, suggesting that longer context does not automatically translate into better predictive signal.

\section{Methodologies}
\subsection{Study cohort and data source}
\begin{figure}
	\centering
	\includegraphics[width=1.0\linewidth]{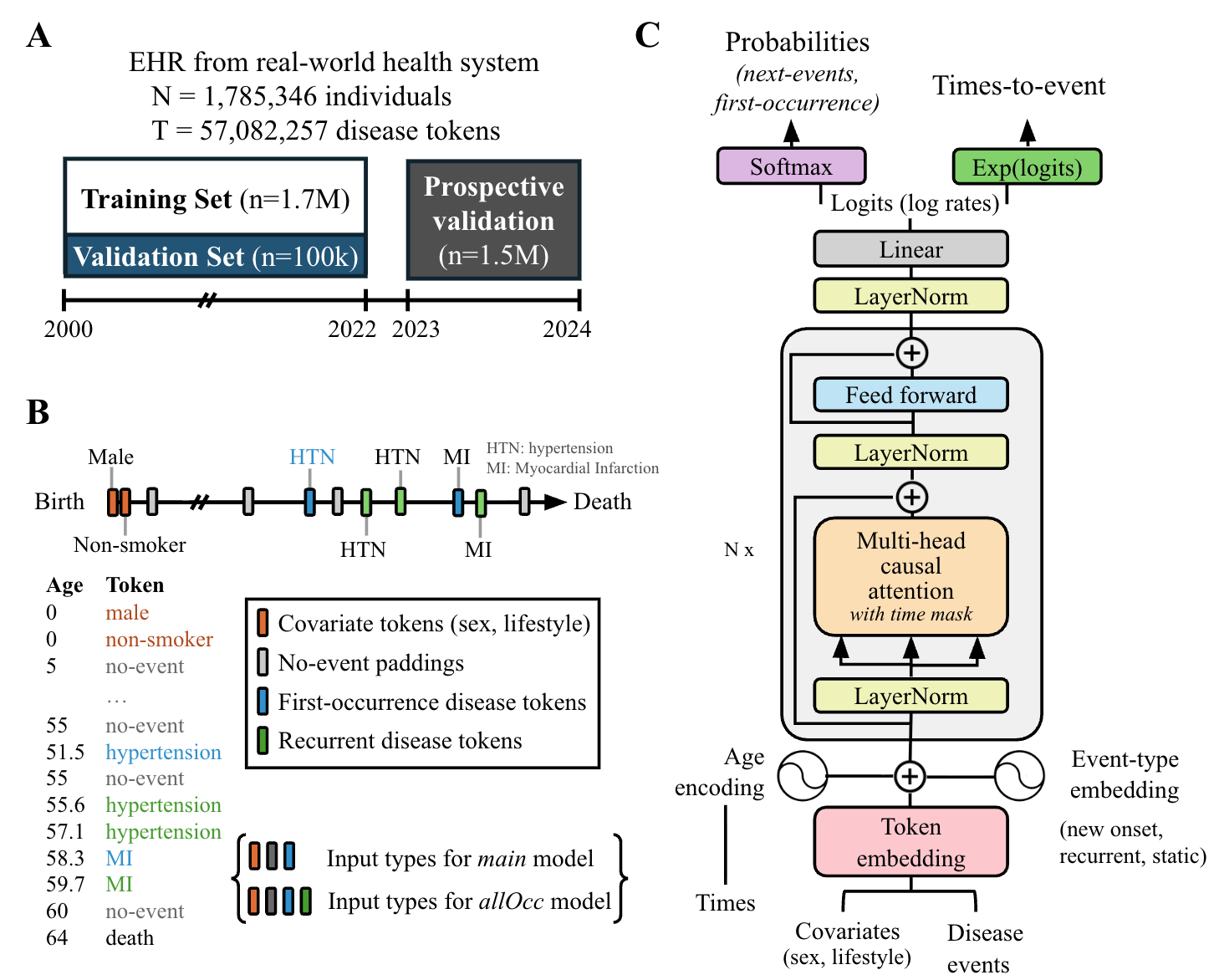}
	\caption{Overview of (A) data, (B) input sequence formatting, and (C) adapted Delphi-style GPT architecture}
	\label{Fig1}
\end{figure}

We used longitudinal structured EHR data spanning 2000--2024 from Mass General Brigham (MGB), a large non-profit integrated health system covering 11 hospitals and 200+ outpatient clinics \cite{desaiBroadeningReachFDASentinel2021} (see Appendix \ref{A_data}). The dataset included diagnosis records (International Classification of Diseases, ICD-9 and ICD-10), basic demographic information, smoking and alcohol status measurements, and date of death records. For model development, patient histories were truncated at Dec-31-2022, leaving 1,785,346 patients with valid non-empty histories. We held out 99,903 patients for validation and used the remaining 1,685,443 patients for training. We also conducted a prospective evaluation to approximate real-world deployment. Among the 1,524,783 patients alive at the end of 2022, we used records up to Dec-31-2022 as historical context for model inputs, 2023 as a temporal gap period, and first-occurrence disease events in 2024 as prospective evaluation targets (Figure~\ref{Fig1}A).

\subsection{Disease trajectory representation}
We transformed each patient’s structured EHR history into a time-ordered sequence of ICD diagnosis events (Figure~\ref{Fig1}B, Appendix \ref{A_data}). For the primary model, repeated diagnoses of the same ICD category were collapsed to the earliest recorded occurrence, so each disease appeared at most once per patient timeline. Diagnosis tokens were paired with the patient’s age in days at the time of diagnosis. Death events were appended at the recorded age of death. Sex, smoking and alcohol status were included as time-fixed covariate inputs but were not used as prediction targets (Appendix \ref{A_data}). Following Delphi, ``no-event'' placeholder tokens were inserted at random intervals to represent elapsed time without recorded events, allowing the model to capture changes in baseline risk across age. The resulting token vocabulary consisted of 1,375 unique ICD categories together with tokens for sex, smoking status, alcohol status, death, and ``no-event'' states.

\subsection{Adapted Delphi-style transformer model}
DT-transformer used a Delphi-style \cite{shmatkoLearningNaturalHistory2025} generative disease trajectory framework for predicting future events from a patient's prior clinical history, adapted to the MGB EHR data (Figure~\ref{Fig1}C). Because EHR events occur on a continuous time axis, this framework represents temporal position using a sinusoidal encoding of continuous age rather than standard positional encoding. At each prediction position, the model produced two outputs: a next-event distribution over disease and death tokens, and a time-to-event prediction modeled through an exponential waiting-time distribution. Together, these outputs allowed the model to estimate future event rates for each disease category. Causal attention masks and same-time masking ensured that predictions were based only on prior events. The training objective combined cross-entropy loss for next-event prediction with the negative log-likelihood of the observed time to next event. No-event tokens and the time-fixed covariate tokens were included as input context but excluded from the prediction loss. The primary model used a fixed-length context window of 93 tokens, consisting of three static covariate tokens and up to 90 prior diagnosis tokens. The model had approximately 2.2 million parameters and was trained with an initial learning rate of $6\times10^{-4}$. Additional architectural and optimization details are provided in Appendix \ref{A_model}.

\subsection{Model evaluation}
\subsubsection{Held-out evaluation}
\label{sec:eval_held.out}
We evaluated whether the model could predict future first-occurrence diagnosis events across a broad disease spectrum in the held-out validation set. Model performance was assessed using age- and sex-stratified areas under the receiver operating characteristic curve (AUCs), across multiple prediction horizons (next-event, 6 months, 1 year, 2 years, and 3 years). Following Delphi, we discarded some ICD outcomes and sparse age-sex strata to reduce instability (Appendix \ref{A_eval_methods}). For each disease, case patients were those who later developed a first recorded occurrence of the target disease and controls were patients who did not. Prediction samples were stratified by sex and 5-year age bins from 20 to 90 years, and disease-specific AUCs were computed within each age-sex stratum and averaged across strata. We also compared performance against an age and sex-based epidemiological baseline for each disease category.

\begin{table}[h]
\centering
\caption{Dataset characteristics by split. Train and Validation are summarized through 2022-12-31; Prospective test summarizes 2024 first-occurrence targets among patients alive at 2022-12-31.}
\label{tab:model_ready_stats}

\resizebox{\linewidth}{!}{
\begin{tabular}{lrrr}
\toprule
\textbf{Statistic} & \textbf{Train} & \textbf{Validation} & \textbf{Prospective test} \\
\midrule
Patients & 1,685,443 & 99,903 & 1,524,783 \\
Diagnosis events & 53,692,361 & 3,161,877 & 2,659,377 \\
Age$^{*}$, median (IQR) & 68.0 (43.4--79.3) & 68.1 (43.5--79.3) & 65.4 (39.6--77.1) \\
History length$^{\dagger}$, median (IQR) & 6.1 (0.9--13.4) & 6.1 (0.9--13.4) & --- \\
Diagnoses/patient$^{\ddagger}$, median (IQR) & 18.0 (6.0--44.0) & 18.0 (6.0--44.0) & --- \\
Female & 56.3\% & 56.1\% & 57.3\% \\
\bottomrule
\end{tabular}
}

\vspace{0.5em}
\begin{minipage}{0.95\linewidth}
\scriptsize
$^{*}$ Age defined as of 2022-12-31, or date of death if earlier. \\
$^{\dagger}$ Time from first to last recorded disease event per patient. \\
$^{\ddagger}$ Number of unique first-occurrence ICD-10 codes per patient within the model vocabulary. \\
$^{\dagger,\ddagger}$ Not applicable to the prospective test column, which summarizes the first-occurrence targets in 2024 only.
\end{minipage}
\end{table}

\subsubsection{Prospective evaluation}
Among 1,524,783 patients in the prospective evaluation cohort, disease events through Dec-31-2022 were used as model context for prediction of first-occurrence events recorded from Jan-1-2024 to Dec-31-2024. Prospective AUC was computed using the same framework described in \ref{sec:eval_held.out}. Calibration was also assessed by comparing predicted and observed disease incidence rates.

\subsection{Recurrent diagnosis extension}
The primary Delphi-style model represents each condition using only its first recorded occurrence, discarding information from repeated diagnoses. We hypothesized that retaining recurrence signals as input context could provide additional information about longitudinal trajectories and evolving comorbidity patterns, potentially improving prediction of future disease onset. To test this, we trained a separate all-occurrence model variant (\textit{allOcc}) in which subsequent records of previously diagnosed conditions were retained as input context. Input diagnosis events were labeled as new-onset or recurrent via an additive flag embedding, and the context window was extended to accommodate longer input sequences (Appendix \ref{A_model}). Recurrent events were excluded from the prediction loss, so the model remained trained to predict future first-occurrence events rather than repeated events.

\section{Experiments and Results}
\subsection{Next-event prediction across the disease spectrum}
\begin{figure}[!t]
	\centering
	\includegraphics[width=\columnwidth]{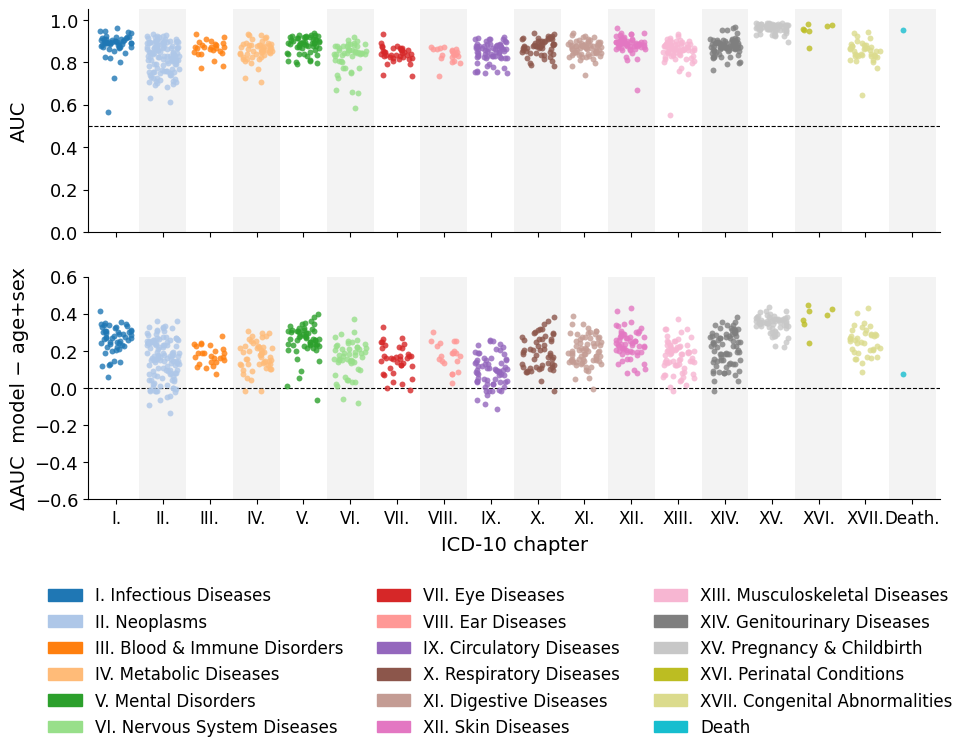}
	\caption{Average age- and sex-stratified AUC values for each disease within held-out validation set, grouped by ICD chapter. Upper panel shows absolute AUCs and lower panel shows difference in AUCs relative to demographic baseline.}
	\label{fig2}
\end{figure}

DT-Transformer predicted next-event outcomes across the range of ICD disease categories. The median age- and sex-stratified AUC was 0.871 (IQR, 0.837--0.898) across all 896 diseases in the validation data (Figure~\ref{fig2}). All disease-specific AUCs exceeded 0.5, indicating that predictive signal was retained across the full set of evaluable outcomes. Relative to the age and sex-based baselines, our model showed a median AUC lift of +0.214 (0.871 vs 0.657), out-performing demographic baselines for 96\% of diseases.

\subsection{Performance across prediction horizons}
\begin{figure}[!t]
  \begin{center}
	\centerline{\includegraphics[width=\columnwidth]{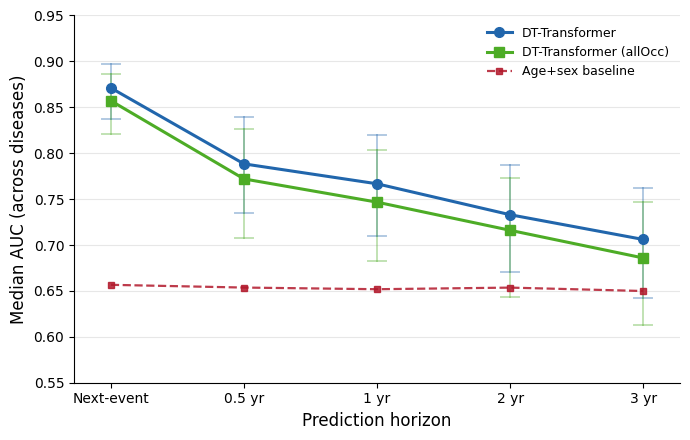}}
	\caption{Median age- and sex-stratified AUC values across prediction horizons in held-out validation. Compares main DT-Transformer model, \textit{allOcc} variant, and demographic baseline.}
	\label{fig.time.horizons}
  \end{center}
\end{figure}

To assess whether predictive signal extended beyond the next-event setting (median 42 days), we evaluated model performance at five horizons ranging from next-event to three years ahead (Figure~\ref{fig.time.horizons}). The model maintained discrimination above the demographic baseline across all horizons, with median AUC remaining above 0.75 within one year and above 0.70 at three years. The demographic baseline was nearly flat across all horizons (0.650--0.657), suggesting that the decline in DT-Transformer AUC reflects attenuation of the diagnostic history signal. The widening interquartile range at longer horizons (0.060 next-event, 0.120 at three years) indicates that long-range predictability becomes increasingly disease-specific.

\subsection{Prospective evaluation performance}
In the prospective evaluation, the model achieved a median AUC of 0.713 across 898 evaluable diseases and outperformed the demographic baseline in 80\% of diseases, demonstrating that discrimination performance generalized to a future time period not encountered in training (Appendix \ref{A_results_prospective}). In calibration analyses, predicted annual incidence rates tracked observed incidence across several orders of magnitude (Appendix \ref{A_results_prospective}), suggesting that model-derived rates retain correspondence to the observed incidence scale.

\begin{table}[t]
\centering
\footnotesize
\setlength{\tabcolsep}{3pt}

\caption{Prospective evaluation performance in 2024 by ICD-10 chapter, reporting total incident first-occurrence cases, mean model AUC, and mean age+sex baseline AUC.}
\label{tab:longit_chapter}

\begin{tabular}{@{}p{4.6cm}rrr@{}}
\toprule
\textbf{ICD-10 Chapter} & \textbf{\makecell{\# of \\Patients}} & \textbf{\makecell{Mean\\AUC}} & \textbf{\makecell{Baseline\\AUC}} \\
\midrule
I. Infectious Diseases & 36,816 & 0.731 & 0.642 \\
II. Neoplasms & 69,982 & 0.657 & 0.641 \\
III. Blood \& Immune Disorders & 63,924 & 0.722 & 0.622 \\
IV. Metabolic Diseases & 119,738 & 0.726 & 0.634 \\
V. Mental Disorders & 73,153 & 0.731 & 0.678 \\
VI. Nervous System Diseases & 76,906 & 0.708 & 0.630 \\
VII. Eye Diseases & 68,942 & 0.716 & 0.613 \\
VIII. Ear Diseases & 38,910 & 0.709 & 0.612 \\
IX. Circulatory Diseases & 167,787 & 0.720 & 0.655 \\
X. Respiratory Diseases & 94,627 & 0.730 & 0.631 \\
XI. Digestive Diseases & 142,450 & 0.709 & 0.626 \\
XII. Skin Diseases & 95,118 & 0.747 & 0.631 \\
XIII. Musculoskeletal Diseases & 202,918 & 0.724 & 0.642 \\
XIV. Genitourinary Diseases & 110,742 & 0.720 & 0.669 \\
XV. Pregnancy \& Childbirth & 32,861 & 0.672 & 0.704 \\
XVI. Perinatal Conditions & 161 & 0.688 & 0.605 \\
XVII. Congenital Abnormalities & 5,893 & 0.670 & 0.630 \\
Death & 7,674 & 0.693 & 0.716 \\
\bottomrule
\end{tabular}
\end{table}

\subsection{Recurrent diagnosis model extension}
The \textit{allOcc} model consistently underperformed the main first-occurrence model at every horizon in the held-out validation (next-event, AUC -0.014; 0.5 year, -0.016; 1 year, -0.020; see Appendix \ref{A_results_allOcc}), whereas the difference in the prospective setting was negligible (+0.005). Thus, simply providing the model with longer histories by retaining recurrent events did not improve first-occurrence prediction in our setting. Examination of the \textit{allOcc} training sequences shows that short-term recurrent events dominated the input context, likely representing repeated documentation within ongoing care episodes rather than biologically distinct events (Appendix \ref{A_results_allOcc}).

\section{Discussion}
We trained a transformer framework on longitudinal structured EHR data from MGB, and evaluated it across nearly 900 disease predictions. The framework retained a meaningful predictive signal in this new setting. Compared to Delphi-2M \cite{shmatkoLearningNaturalHistory2025}, DT-Transformer demonstrated stronger near-term discrimination followed by faster long-horizon decay. While direct comparisons are complicated by the different underlying populations (Appendix \ref{A_results}), differences in data density and longitudinal coverage may explain these results. Unlike sparser, biobank-style databases of Delphi-2M, EHR sequences are organized around clinical encounters and tend to contain richer recent history, prioritizing short-horizon forecasting. A second issue is longitudinal completeness in the fragmented US system: while MGB patient timelines extend beyond a single hospital, events that occur outside the MGB network remain unobserved \cite{getzenMiningEquitableHealth2023, karImpactElectronicHealth2023a}. This difference in observability may be particularly relevant for longer-term prediction, where multi-year co-occurrence trajectories provide critical context.

This study has the following limitations. The time encoding with age embeddings and no-event paddings may be suboptimal for real-world EHR data, where events are more tightly spaced and long-term observability and censoring are more pronounced. More explicit interval-aware temporal encoding \cite{rencZeroShotHealth2024a} may provide a better match. Additionally, although our data spans MGB and captures events across affiliated clinics and facilities, events occurring outside the MGB network remain unobserved, which may be particularly consequential for longer-horizon prediction.

Future work could extend the training objective to explore whether additional supervision beyond immediate next-event prediction can better support longer-horizon forecasting. For example, use of multi-token prediction objectives from NLP \cite{gloeckleBetterFasterLarge2024}. On the input side, extensions could incorporate additional structured (e.g., procedures and medications), unstructured (e.g., free-text notes) EHR sources \cite{wu_bridge_2025, liuMultimodalDataMatters2022, shaoScalableMedication2025, guScalableInformation2025}, administrative claims data, as well as external knowledge \cite{wuLargeLanguageModels2024}. On the output side, and of particular relevance for real-world deployment, model extensions could move beyond first-occurrence prediction to capture a broader range of clinically important trajectories, including chronic disease flare-ups and acute hospitalization events.

\section{Conclusion}
DT-Transformer demonstrates that a GPT-based trajectory modeling framework can achieve strong next-event prediction when trained on structured EHRs from MGB. Whether such models can ultimately inform clinical decision-making or system-level forecasting remains an open question, requiring further work on model development and alternative evaluation paradigms.

\section*{Impact Statement}
This work advances foundation models for modeling disease trajectories using real-world EHR data. The potential societal impacts are consistent with prior work on clinical prediction models, including improved risk stratification and healthcare planning. At the same time, such models may inherit biases present in observational healthcare data and should be carefully evaluated for fairness, reliability, and appropriate use before deployment.

\bibliography{references}
\bibliographystyle{plainnat}

\clearpage
\appendix
\onecolumn
\section{Appendix}

\renewcommand{\thefigure}{A\arabic{figure}}
\setcounter{figure}{0}

\renewcommand{\thetable}{A\arabic{table}}
\setcounter{table}{0}

\subsection{Dataset, data processing, and input formats}
\label{A_data}
We used data from a 1.7 million patient sample drawn from Mass General Brigham (MGB), a large integrated health system serving a broader population \cite{desaiBroadeningReachFDASentinel2021}. Structured diagnoses were obtained from the longitudinal EHR covering all encounters across the network, which comprises 11 hospitals and several hundred affiliated outpatient clinics.

Patient sequences were constructed for individuals with at least one diagnosis record during the study period. We required a non-missing date of birth and known sex, and excluded patients with implausible temporal ordering, such as diagnosis events occurring before birth or after the recorded date of death. The diagnosis records were standardized by mapping the ICD-9 codes to ICD-10 using the 2018 CMS General Equivalence Mappings crosswalk, followed by a truncation to 3-character disease categories. Disease categories recorded in fewer than 1,000 patients were excluded to reduce sparsity and improve the stability of downstream model training and evaluation; this threshold removed 27.7\% of the disease vocabulary while accounting for only 0.2\% of diagnosis events.

In addition to diagnosis events, we incorporated sex, smoking status, and alcohol status as time-fixed input covariates. These variables were used only as historical context and were not included as prediction targets. Smoking and alcohol status were defined using the most recent valid measurement before the patient’s first diagnosis record; if no prior valid measurement was available, the earliest valid measurement after the first diagnosis record was used. Missing or unclassifiable values were assigned to an Unknown category. All covariate tokens were inserted at age zero so that they were available as static context throughout each patient sequence.

\subsection{Model details}
\label{A_model}

The model uses a fixed-length context window. For the main (first-occurrence) model, the input length was set to 93 tokens, consisting of three static covariate tokens and up to 90 diagnosis tokens. This diagnosis context length was chosen to approximately cover the 90th percentile of patient sequence lengths in the training data.

For the \textit{allOcc} model, the context window was increased to 445 tokens to accommodate the longer sequences generated by retaining recurrent diagnosis events. This length was similarly selected based on the empirical sequence length distribution in the all-occurrence training data. To distinguish incident and recurrent diagnosis tokens, the \textit{allOcc} model included an additional learned flag embedding. Each input token was assigned one of three flags: static or no-event token, new-onset disease token, or recurrent disease token. The corresponding flag embedding was added to the token and age embeddings before the transformer blocks.

The model used 12 transformer layers, 12 attention heads, and an embedding dimension of 120, with approximately 2.2 million parameters in total.

Training used an initial learning rate of $6\times10^{-4}$ with 1,000 warmup iterations, followed by cosine decay to a minimum learning rate of $6\times10^{-5}$, and weight decay of 0.2. The final checkpoint was selected based on validation loss.

In the \textit{allOcc} model, recurrent diagnosis events were retained as input context but masked from the prediction loss. Thus, both the primary and allOcc models were trained and evaluated for first-occurrence prediction, making their AUCs directly comparable.

We did not perform extensive hyperparameter tuning on the held-out validation set; instead, we used a fixed model configuration across experiments.

\subsection{Evaluation methodology}
\label{A_eval_methods}

Following the Delphi framework, we excluded non-disease ICD chapters when calculating disease-level AUCs, including symptoms, injuries, external causes, health-status factors, and unknown categories.

To ensure stable estimation of model performance in the held-out evaluation, we applied additional filtering criteria. Age-by-sex strata with fewer than six disease cases were excluded, and disease-level AUC was reported only for diseases with at least two valid age-by-sex strata. Prediction inputs were also restricted to person-time between ages 20 and 90 years due to limited event counts outside this range. For the prospective evaluation, we used a modified filtering strategy because outcomes were restricted to events within a single calendar year, 2024. We required each disease to have at least 25 total cases in the prospective evaluation set and required both case and control samples to be present for valid AUC estimation.

For each disease, case patients were those who later developed a first recorded occurrence of the target disease and controls were patients who did not. For cases, predictions were made before the first recorded occurrence; for controls, prediction points were sampled from available clinical histories. Patients with prior history of the target disease were excluded from the case-control set for that disease to keep evaluation focused on first recorded (incident) occurrence. Prediction samples were stratified by sex and 5-year age bins from 20 to 90 years. Disease-specific AUCs were computed within each valid age-sex stratum and averaged across strata. Across-disease performance was summarized using the median and interquartile range of disease-level AUCs.

The demographic-based epidemiological baseline assigned disease risk based on observed incidence within matched age-sex strata and was evaluated using the same case-control samples as DT-Transformer.

\subsection{Extended results}
\label{A_results}

\subsubsection{Extended results from held-out validation for main model}
\label{A_results_held.out}

This table provides a detailed summary of model performance across prediction horizons in the held-out validation set, corresponding to the horizon analysis in the main text.

\begin{table}[h]
\centering
\caption{Summary of DT-Transformer and age-sex baseline performance across prediction horizons in the held-out validation set.}
\label{tab:horizon_sweep}

\begin{tabular}{lcccc}
\toprule
\textbf{Horizon} & \textbf{Median AUC} & \textbf{IQR}
& \textbf{Baseline AUC} & $\mathbf{\Delta}$ \textbf{AUC} \\
\midrule
Next-event & 0.871 & (0.837--0.898) & 0.657 & $+$0.214 \\
0.5 yr     & 0.788 & (0.735--0.840) & 0.654 & $+$0.135 \\
1 yr       & 0.767 & (0.710--0.820) & 0.652 & $+$0.115 \\
2 yr       & 0.733 & (0.671--0.787) & 0.654 & $+$0.079 \\
3 yr       & 0.706 & (0.642--0.762) & 0.650 & $+$0.056 \\
\bottomrule
\end{tabular}

\end{table}

\subsubsection{Extended results from prospective validation for main model}
\label{A_results_prospective}
\begin{figure}[htbp]
	\centering
	\includegraphics[width=0.75\linewidth]{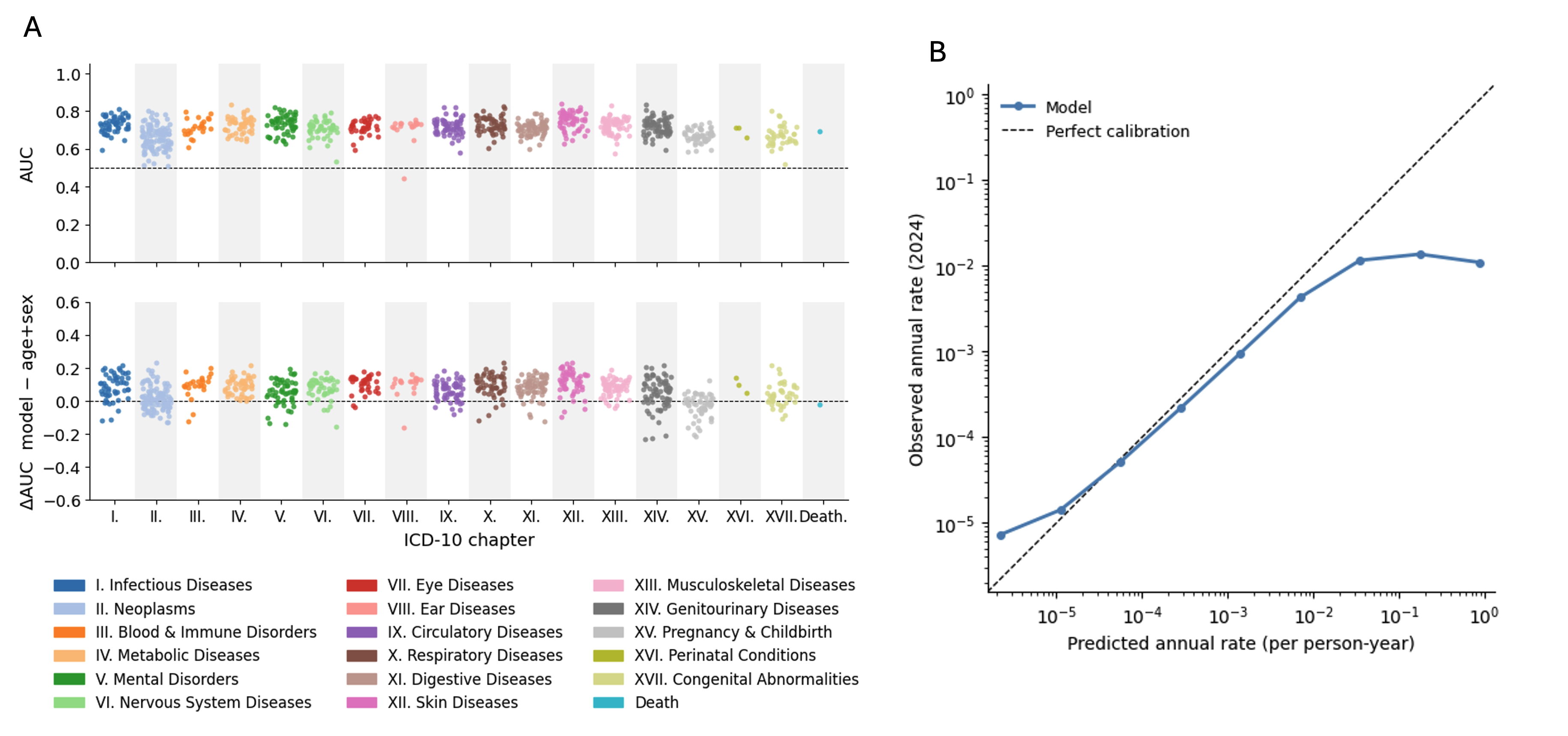}
	\caption{Prospective longitudinal evaluation of the main model. (A) Disease-level AUC and lift over the age--sex baseline across ICD-10 chapters. (B) Calibration of predicted annual incidence rates for 2024 events.}
	\label{fig:prospective_extended}
\end{figure}

Figure~\ref{fig:prospective_extended} shows the disease-level results underlying the prospective evaluation summarized in Section 4.3. The left panels display prospective AUC and $\Delta$AUC relative to the age-sex baseline for individual diseases, grouped by ICD-10 chapter. The right panel shows the corresponding calibration curve used to assess agreement between predicted and observed annual incidence rates in 2024.

\subsubsection{Extended subgroup analysis for main model}
\label{A_subgroup}
\begin{figure}[htbp]
	\centering
	\includegraphics[width=\linewidth]{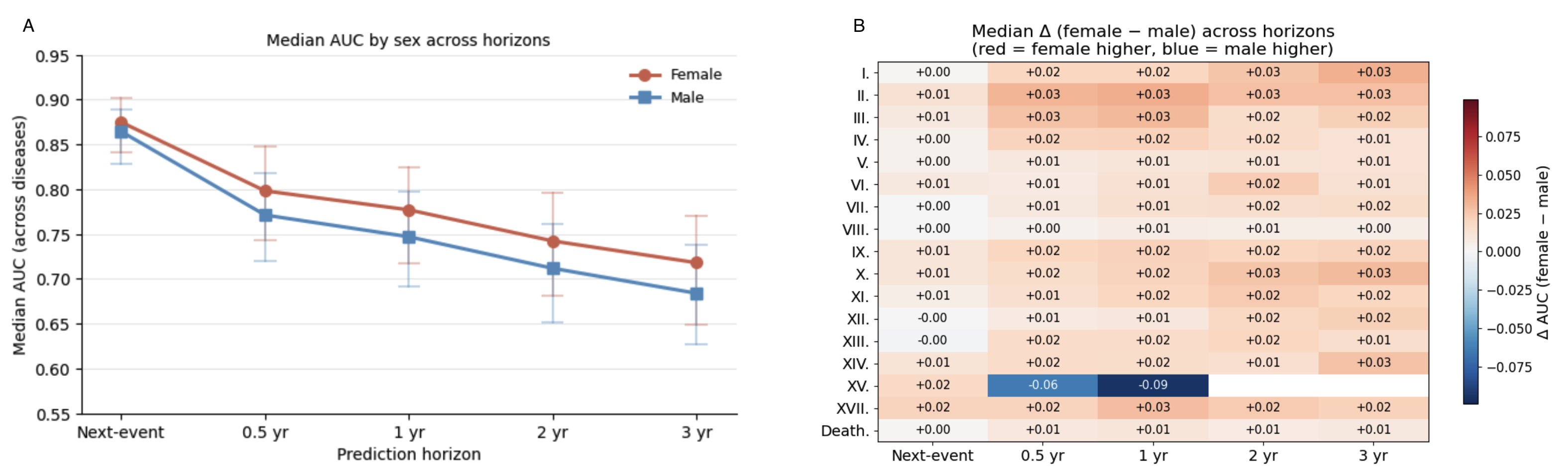}
	\caption{Sex-stratified prediction performance in the held-out validation set. (A) Median AUC across diseases by sex and prediction horizon; error bars indicate interquartile range. (B) Median AUC difference (female $-$ male) across ICD chapters and prediction horizons, computed as the median across diseases within each chapter.}
	\label{fig:sex_combined}
\end{figure}

\textbf{Performance by sex.} We stratified model performance by sex. As shown in Figure~\ref{fig:sex_combined}, prediction performance is consistently higher in females than in males across all prediction horizons. While the difference is relatively small for next-event prediction, the gap becomes wider at longer horizons. To examine whether this difference is driven by a small subset of diseases, we further analyzed performance across ICD chapters (Figure~\ref{fig:sex_combined}). Female patients exhibit higher AUC across most chapters and horizons, suggesting that the observed performance gap is systematic rather than confined to specific disease groups.

\textbf{Performance by age.} As shown in Figure~\ref{fig:subgroup_age}, model performance was broadly comparable across age groups, with AUCs remaining within a similar range across the evaluated age bins. This suggests that the model maintains relatively stable discrimination across patient age groups in the prospective setting.

\begin{center}
	\includegraphics[width=0.62\linewidth]{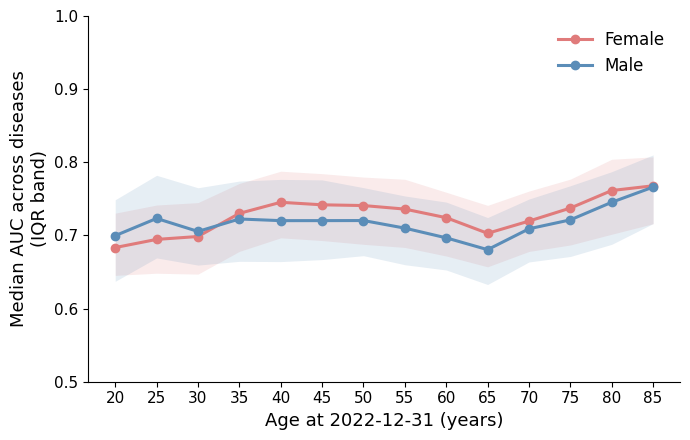}
	\captionof{figure}{Prospective model performance by patient age at the 2022 cutoff. Each point shows the median AUC across diseases within an age group.}
	\label{fig:subgroup_age}
\end{center}

\subsubsection{Extended results from \textit{allOcc} model}
\label{A_results_allOcc}

We provide additional results for the \textit{allOcc} model and compare it with the primary model. Table~\ref{tab:auc_comparison} reports the median disease-level AUC across prediction horizons for both models.

\begin{figure}[htbp]
	\centering
	\includegraphics[width=0.65\linewidth]{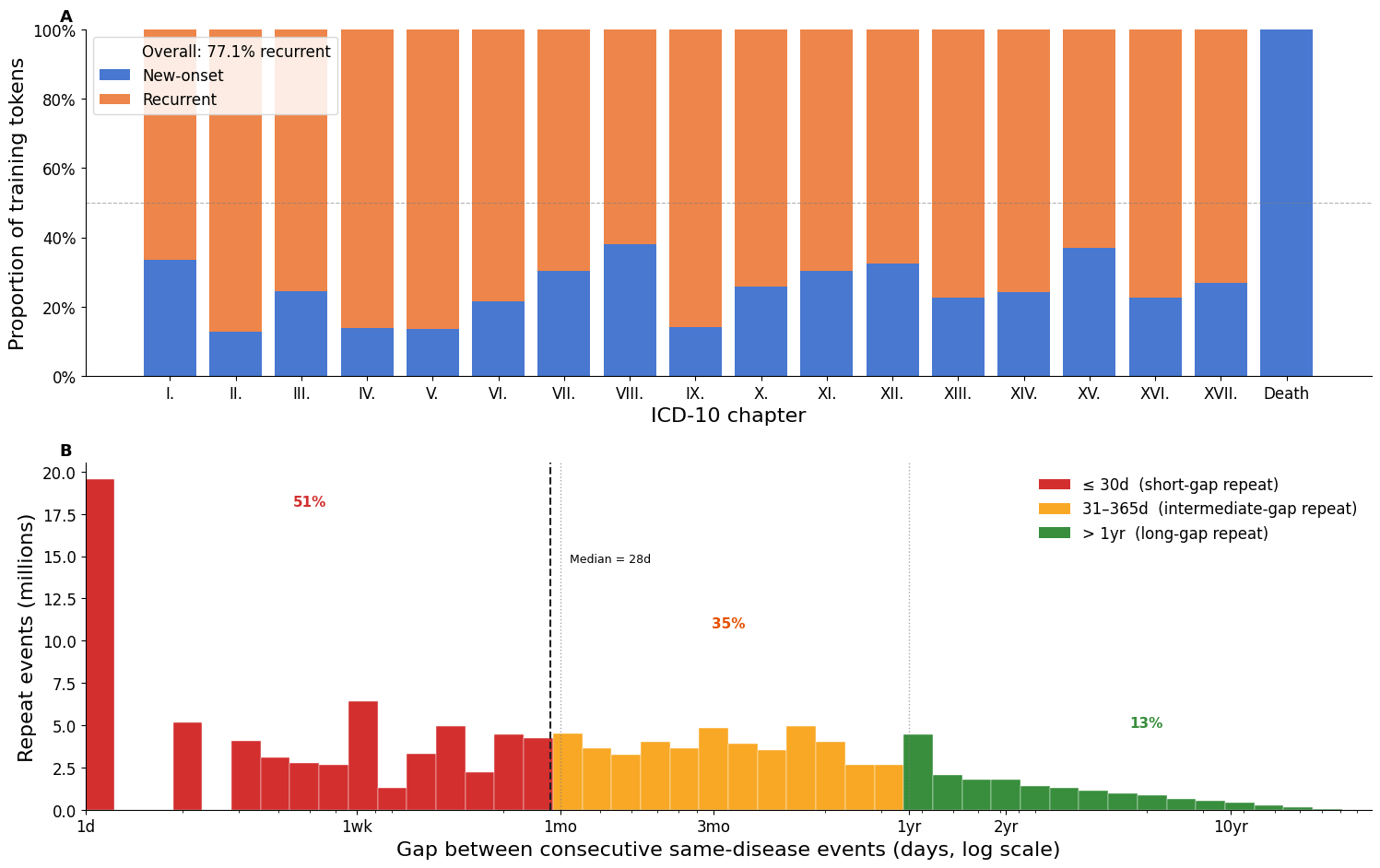}
	\caption{Composition and temporal spacing of recurrent diagnosis tokens in the \textit{allOcc} training sequences. (A) Proportion of disease tokens corresponding to new-onset and recurrent diagnoses across ICD-10 chapters. (B) Distribution of time gaps between consecutive records of the same disease among recurrent events, shown on a log-scaled x-axis.}
	\label{fig:allocc_composition}
\end{figure}

To characterize the input sequences used by \textit{allOcc}, Figure~\ref{fig:allocc_composition} summarizes the contribution and timing of recurrent diagnosis tokens. Panel A shows the proportion of new-onset and recurrent disease tokens by ICD-10 chapter. Overall, recurrent events accounted for 77.1\% of disease tokens in the \textit{allOcc} training sequences. Panel B shows the distribution of gaps between consecutive records of the same disease among recurrent events. Approximately 51\% of recurrent events occurred within 30 days of the immediately preceding same-disease record.

\begin{table}[htbp]
\centering
\caption{Median AUC by prediction horizon and model variant. Retrospective: per-disease median AUC across diseases passing the paper filter ($\geq 6$ cases per age-sex bin, $\geq 2$ valid bins). Prospective: per-disease median AUC across all diseases present in the longitudinal evaluation ($\geq 25$ cases enforced at evaluation time). $\Delta = \text{allOcc} - \text{firstOcc}$; positive favours allOcc.}
\label{tab:auc_comparison}
\begin{tabular}{lccccc}
\toprule
& \multicolumn{2}{c}{\textbf{firstOcc}}
& \multicolumn{2}{c}{\textbf{allOcc}}
& \\
\cmidrule(lr){2-3}\cmidrule(lr){4-5}
\textbf{Horizon} & $N$ & Median AUC & $N$ & Median AUC
& $\Delta$ (allOcc$-$firstOcc) \\
\midrule
\multicolumn{6}{l}{\textit{Held-out set}} \\
\quad Next-event & 896 & 0.871 & 887 & 0.857 & $-$0.014 \\
\quad 0.5 yr     & 890 & 0.788 & 886 & 0.772 & $-$0.016 \\
\quad 1 yr       & 891 & 0.767 & 877 & 0.747 & $-$0.020 \\
\quad 2 yr       & 886 & 0.733 & 876 & 0.716 & $-$0.017 \\
\quad 3 yr       & 874 & 0.706 & 873 & 0.686 & $-$0.020 \\
\midrule
\multicolumn{6}{l}{\textit{Prospective set}} \\
\quad 2024       & 898 & 0.713 & 898 & 0.718 & $+$0.005 \\
\bottomrule
\end{tabular}
\end{table}
\FloatBarrier

\end{document}